\documentclass[lettersize,journal]{IEEEtran}
\usepackage{algorithmicx}
\usepackage{algorithm}
\usepackage{array}
\usepackage[caption=false,font=normalsize,labelfont=sf,textfont=sf]{subfig}
\usepackage{textcomp}
\usepackage{stfloats}
\usepackage{url}
\usepackage{verbatim}
\usepackage{graphicx}
\usepackage{cite}
\usepackage{amsmath,amssymb,amsfonts}
\usepackage{xcolor}
\usepackage{bm}
\usepackage{amsmath}
\usepackage{amssymb}
\usepackage{amsthm}
\usepackage{mathrsfs}
\usepackage{enumerate}
\usepackage{multirow}
\usepackage{threeparttable}
\usepackage[square, comma, sort&compress, numbers]{natbib}
\usepackage[pagebackref=false,breaklinks=true,letterpaper=true,colorlinks,bookmarks=false]{hyperref}
\usepackage{times}
\usepackage{breakurl}
\usepackage{verbatim}
\usepackage{bbding}
\usepackage{algpseudocode}
\usepackage{lettrine}
\usepackage{float}
\usepackage{tabularx} 
\usepackage{makecell}  
\usepackage{booktabs}   
\begin{document}

\title{PQ-DAF: Pose-driven Quality-controlled Data Augmentation for Data-scarce Driver Distraction Detection}

\author{
    Haibin Sun$^{a}$, Xinghui Song$^{a,*}$%
    \thanks{*Corresponding author. Email addresses: \texttt{sdustsun@163.com} (Haibin Sun), 
    \texttt{xinghuisong7026@163.com} (Xinghui Song).}%
    \thanks{$^{a}$College of Computer Science and Engineering, Shandong University of Science and Technology, Qingdao, 266590, China.}%
}



\maketitle

\begin{abstract}

Driver distraction detection is essential for improving traffic safety and reducing road accidents. However, existing models often suffer from degraded generalization when deployed in real-world scenarios. This limitation primarily arises from the few-shot learning challenge caused by the high cost of data annotation in practical environments, as well as the substantial domain shift between training datasets and target deployment conditions. To address these issues, we propose a Pose-driven Quality-controlled Data Augmentation Framework (PQ-DAF) that leverages a vision-language model for sample filtering to cost-effectively expand training data and enhance cross-domain robustness. Specifically, we employ a Progressive Conditional Diffusion Model (PCDMs) to accurately capture key driver pose features and synthesize diverse training examples. A sample quality assessment module, built upon the CogVLM vision-language model, is then introduced to filter out low-quality synthetic samples based on a confidence threshold, ensuring the reliability of the augmented dataset. Extensive experiments demonstrate that PQ-DAF substantially improves performance in few-shot driver distraction detection, achieving significant gains in model generalization under data-scarce conditions.
\end{abstract}


\section{Introduction}
\IEEEPARstart{D}{river} distraction detection plays a vital role in enhancing traffic safety and reducing accident risk. In recent years, deep learning-based approaches have become the mainstream for this task, achieving remarkable results. However, due to the domain shift between training data and real-world deployment scenarios, these models often suffer substantial performance degradation in practice. To mitigate such data shift, collecting new datasets at deployment is an intuitive solution. Yet, data collection and annotation are costly and labor-intensive, making it crucial to improve model robustness under few-shot conditions.

Existing research can be broadly divided into two directions. The first focuses on designing lightweight and efficient classification models optimized for deployment on resource-constrained edge devices. For instance, MTNet integrates a multi-dimensional adaptive feature extraction module with lightweight feature fusion, reducing parameters by 37\% on the LDDB benchmark while maintaining 96.2\% accuracy, albeit with an 8.3\% drop for small targets such as hand movements~\cite{zhu2023driver}. Similarly, RES-SE-CNN combines residual networks with channel attention, achieving 97.28\% accuracy while significantly lowering memory usage, making it suitable for in-vehicle applications~\cite{lei2025intelligent}. Nevertheless, these methods typically require large-scale labeled datasets and remain vulnerable to domain shifts, limiting their generalization.

The second line of work addresses the few-shot problem by mitigating domain shift through data augmentation or synthetic data generation~\cite{cronje2017training}. In other domains, techniques such as image transformation, style transfer, and GAN-based generation have been applied to expand training sets and improve robustness, yet they remain underexplored for driver distraction detection~\cite{zhu2024analysis, ugli2022transfer}. For example, Wang et al.~\cite{wang2020data} achieved 96.97\% accuracy on the StateFarm dataset by integrating region detection with image enhancement, while Hasan et al.~\cite{hasan2024vision} applied a Sim2Real strategy for cross-domain adaptation. Despite their potential, these methods often suffer from unstable sample quality, high labeling costs, and low reliability of synthetic data, which limit their real-world applicability.

To address these challenges, we propose a {Pose-driven Quality-controlled Data Augmentation Framework (PQ-DAF)} tailored for few-shot driver distraction detection. First, driver pose information is extracted via DWpose and used as a conditional input to Progressive Conditional Diffusion Models (PCDMs)~\cite{shen2023advancing} for generating diverse, high-fidelity training samples that preserve structural consistency. Next, the vision-language model CogVLM~\cite{wang2024cogvlm} serves as an automatic quality evaluator, filtering out low-confidence synthetic samples based on a predefined threshold, thus ensuring the reliability of augmented data. This approach eliminates manual annotation, enables cost-effective dataset expansion, and enhances cross-domain generalization. Additionally, we investigate the effect of varying the ratio of real to synthetic data.
The main contributions of this work are summarized as follows:
\begin{itemize}
    \item We capture key human pose information using DWpose and employ PCDMs to synthesize diverse, high-fidelity driver behavior images with structural consistency.
    \item We introduce a CogVLM-based automatic quality filtering mechanism to remove low-quality samples, improving the reliability of augmented datasets.
    \item We conduct extensive experiments on multiple real-world driver behavior datasets, demonstrating that our method consistently improves performance in few-shot scenarios.
\end{itemize}

\section{Related Work}
\subsection{Related Driver Distraction Detection Work}
With the increasing prevalence of vehicles and the growing complexity of road traffic, traffic safety issues have become increasingly severe.  According to statistics from the World Health Organization, approximately 1.35 million people die in traffic accidents worldwide each year ~\cite{ge2022review}. Studies indicate that driver distractions are one of the important causes of road traffic accidents. Therefore, detecting driver distraction is of great significance in the field of road safety.

In the early research on driver distraction detection, researchers primarily relied on traditional machine learning methods combined with handcrafted feature extraction. For example, Billah and Rahman et al. ~\cite{tian2013studying} study driver distraction detection using hand, lip, and forehead features and adopt the K-nearest neighbor (KNN) algorithm, which achieves an accuracy of 81.50\% to 81.67\% on their self-built dataset. Although these methods have achieved certain results in specific scenarios, their feature extraction process is complex and inherently limited. With the advancement of deep learning, convolutional neural network (CNN)--based methods have gradually become the mainstream approach~\cite{poon2021driver,khellal2024distracted,qin2021distracted,mittal2023cat}. For example, The lightweight CNN model OLCNNet  ~\cite{carney2015using} achieves 89.53\% accuracy on the StateFarm dataset. In addition, Transformer models are introduced into the behavior recognition field due to their superior ability to model long-range dependencies ~\cite{fang2022driver,li2024lightweight}. The BiRSwinT ~\cite{yang2023birswint} adopts a dual-stream structure based on the Swin Transformer, learning and fusing global and local driver action features to achieve fine-grained driver behavior recognition. However, although these methods achieve promising recognition accuracy on specific datasets, their performance often degrades significantly in data-scarce scenarios due to domain shifts between training data and real-world application environments. This necessitates the collection of new datasets—a process that is both complex and costly due to the challenges of data acquisition and annotation. To enhance model performance in data-scarce scenarios, acquiring authentic and diverse training data is crucial. However, collecting driver behavior data often faces challenges such as high costs, strong privacy concerns, and complex scenarios, making it difficult to obtain comprehensive, high-quality abnormal behavior samples in real-world settings. Therefore, exploring more effective data expansion and augmentation methods—particularly data generation techniques capable of simulating realistic driving environments and behaviors—has become key to improving the performance of driver distraction detection models in practical applications.

\subsection{Pose-Guided Image Generation Related Work}  
Pose-guided image generation has made remarkable progress in recent years, aiming to synthesize high-quality images that conform to a specified target pose. The central idea is to incorporate target pose information into the generation process so that the synthesized result matches both the style of the source image and the desired pose. Early studies formulated this task as a conditional image generation problem using Conditional Generative Adversarial Networks (CGAN)~\cite{mirza2014conditional}, where the source image appearance and target pose serve as conditioning inputs. However, direct conditioning often struggles with complex pose transformations due to inherent misalignment between the source and target poses. To address this challenge, Def-GAN~\cite{siarohin2018deformable} introduced a deformable GAN that models pose changes via a set of local affine transformations, mitigating misalignment issues. Similarly, ADGAN~\cite{gu2021adain} employed a texture encoder to extract style vectors of body parts and applied them to AdaIN residual blocks to generate the final image. These approaches focus on transferring the style of the source image into the target pose by designing architectures tailored for complex pose transformations, yet they still face limitations in modeling highly non-linear deformations.

Recently, diffusion models~\cite{shen2025long, shen2025imaggarment, shen2025imagharmony} have emerged as powerful generative models, producing superior image quality and diversity compared to GAN-based methods. After their success in unconditional synthesis, diffusion models were rapidly extended to conditional settings, including pose guidance. Early work such as PIDM~\cite{bhunia2023person} adapted Stable Diffusion to jointly condition on source and pose images, enabling effective fusion of style and pose cues. MGD~\cite{baldrati2023multimodal} further leveraged multi-modal conditions—pose maps, clothing sketches, and text—to guide generation in latent space. Progressive Conditional Diffusion Models (PCDMS)~\cite{shen2024advancing} advanced this direction by progressively refining pose and style information, while HumanSD~\cite{ju2023humansd} optimized human generation in complex scenes. Recent unified frameworks such as ImagPose~\cite{shen2024imagpose} and task-specific systems like ImagDressing~\cite{shen2025imagdressing} demonstrate that diffusion-based conditioning can flexibly handle both pose and appearance customization, even in multi-stage workflows~\cite{shen2025boosting}.
To date, however, no study has applied pose-guided image generation to driver distraction detection. Given that distraction behaviors are strongly correlated with pose patterns, and that public datasets rarely match real-world driving scenarios, pose-guided generation offers a promising way to simulate diverse driver behaviors. By synthesizing varied pose configurations, it can enrich training datasets and potentially enhance the robustness and accuracy of distraction detection models in practical deployment.

\section{Proposed Method}

\begin{figure}[t]
    \centering
    \includegraphics[width=1\linewidth]{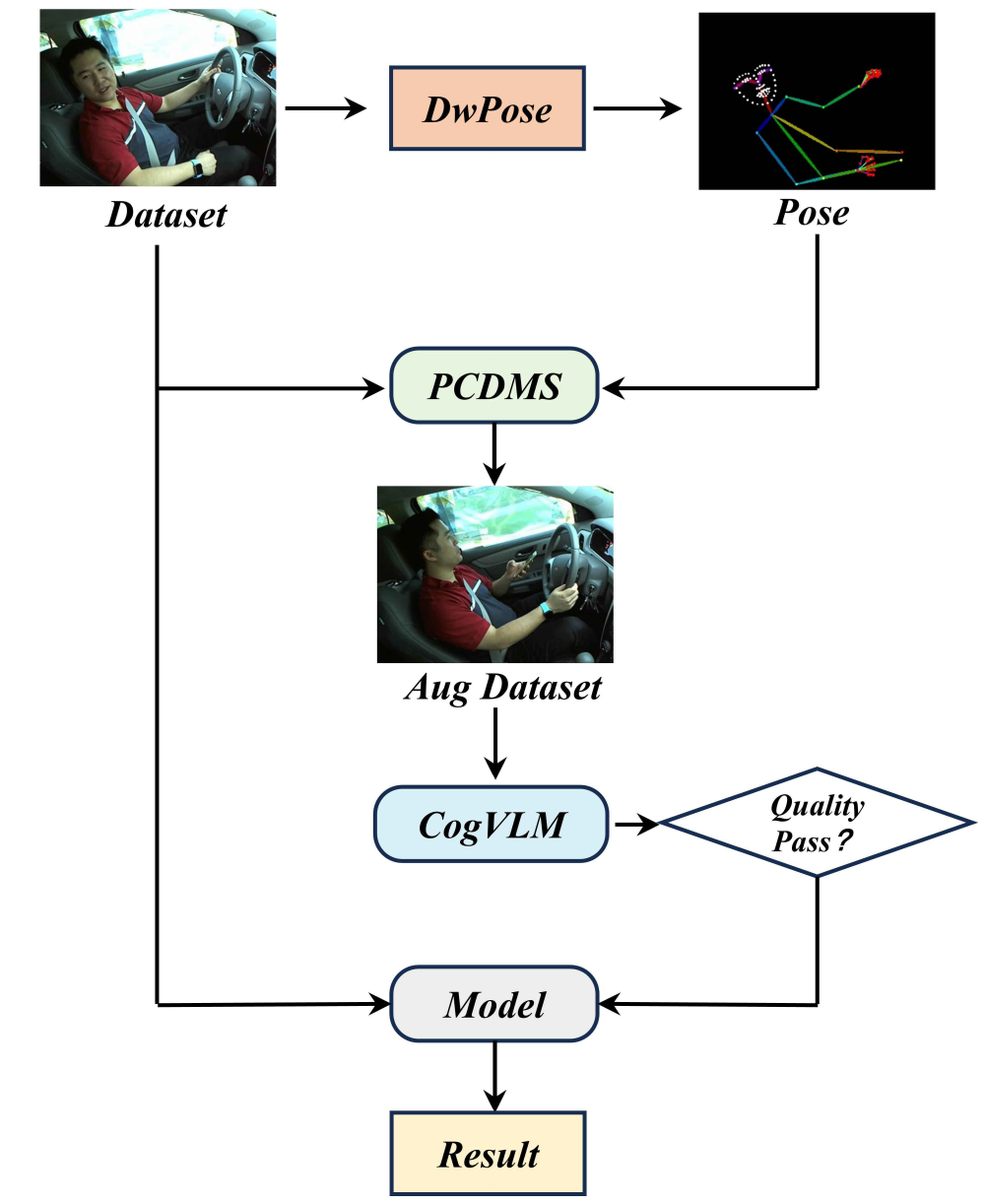}
\caption{ Pose-Driven Quality-Controlled Data Augmentation Framework. Dataset is fed into DwPose for Pose extraction; both are input to PCDMs to generate Aug Dataset. After CogVLM evaluation and Quality Pass? filtering, it is jointly trained with Dataset on Model to output Result. }
\label{fig:1}
\end{figure}
\subsection{Overview}
To tackle the data scarcity problem arising from the high cost of manual annotation in driver distraction detection scenarios,  we propose Pose-Driven Quality-Controlled Data Augmentation Framework (PQ-DAF), As shown in Fig. \ref{fig:1}. First, we extract human keypoint poses from original driving images as structural priors. Then we use progressive conditional diffusion models (PCDMs) to generate diverse, well-structured pseudo-samples that expand the training set. Unlike traditional data augmentation or GAN-based methods, PCDMs offer stronger modeling capacity and greater stability. Under a given pose constraint, they produce semantically clear images that reflect consistent driving actions. 

After sample generation, we introduce an automated quality control mechanism based on the multimodal reasoning of the visual-language model CogVLM. For each pose category, we define a unified semantic prompt and evaluate how well the generated images match that prompt. We compute a confidence score for each image-text pair and discard those below a preset threshold. This filtering step removes noisy samples that could harm training. Finally, we merge the retained high-quality pseudo-samples with real data and retrain the model. By combining structure-guided augmentation with semantic filtering, PQ-DAF effectively augment the training data without additional human annotations, thereby improving model performance in data-scarce scenarios.

\subsection{Pose-Conditioned Image Generation Module}  
We proposes a Pose-Conditioned Image Generation Module that integrates human pose extraction and image generation techniques to synthesize training samples with high semantic consistency and visual realism, effectively addressing the insufficient model generalization capability caused by data scarcity in driver distraction detection tasks. Specifically, DWpose  is first utilized to extract refined skeletal pose maps from both source and target images as structural prior constraints, which are then fed into the second-stage generator of Progressive Conditional Diffusion Models (PCDMs)  to perform latent space conditional diffusion, generating driving behavior images characterized by semantic plausibility and structural coherence.

\begin{figure*}[t]
    \centering
    \includegraphics[width=1\linewidth]{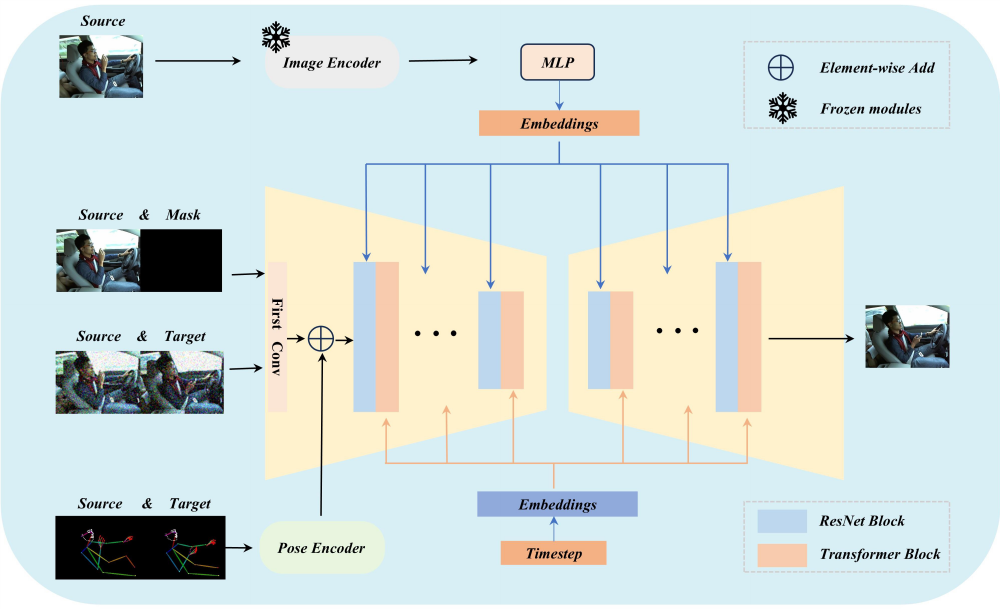}
\caption{Progressive Conditional Diffusion Models}
\label{fig:2}
\end{figure*}

To ensure semantic consistency and effectively model critical regions in generated images, the model input is composed of three conditional branches as shown in Fig. \ref{fig:2}: (1) the source image and its corresponding binary mask; (2) paired source and target images; and (3) source and target pose maps. These image pairs are concatenated along the width dimension to construct a spatially aligned input representation. To prevent ambiguity caused by black pixels, we further introduce a single-channel indicator map (not shown in the figure), where 0 and 1 respectively denote occluded and non-occluded regions, assisting the model in identifying the areas to be inpainted.

For feature encoding, we adopt a frozen DINOv2 \cite{oquab2023dinov2} encoder to extract fine-grained semantic representations from the source image, which are further projected into a latent space via a shallow multilayer perceptron (MLP). The pose maps are processed by a four-layer convolutional Pose Encoder to preserve spatial structure and accurately model the target action. The multi-source information is encoded into three types of embeddings: the fused image embedding $f_{\text{st}}$, the pose structure embedding $p_{\text{st}}$, and the mask-aware embedding $i_{\text{sm}}$, which are jointly fed into the diffusion network.

The backbone UNet consists of alternating ResNet blocks and Transformer blocks, enabling the model to capture both local texture details and global contextual dependencies. During training, the objective is to reconstruct images corrupted with Gaussian noise. The optimization target is to minimize the reconstruction error between the predicted and ground-truth noise, and the loss function $\mathcal{L}$ is defined as:
\begin{equation}
\mathcal{L} = \mathbb{E}_{x_0, \epsilon, t} \left\| \epsilon - \epsilon_\theta (x_t, f_{\text{st}}, p_{\text{st}}, i_{\text{sm}}, t) \right\|_2^2.
\end{equation}
To further improve controllability and generation quality, we adopt the classifier-free guidance strategy during inference. By weighting and fusing the image semantic branch and the pose structural branch, the generation process is effectively guided:

\begin{equation}
\hat{\epsilon}_\theta = w \cdot \epsilon_\theta(x_t, f_{\text{st}}, i_{\text{sm}}, t) + (1 - w) \cdot \epsilon_\theta(x_t, p_{\text{st}}, t), \quad w \in [0, 1].
\end{equation}

Through the above mechanisms, PCDMs demonstrate strong pose controllability and image quality assurance in driver behavior generation tasks. The generated pseudo samples are not only structurally aligned with the target pose but also semantically consistent, effectively mitigating limitations in the size and distribution of the original training dataset.
\subsection{Generated Image Filtering Mechanism}  

Although PCDMs are capable of generating images with high pose consistency, semantic ambiguities may still arise—such as action misclassification (e.g., texting vs. calling) or background artifacts. To ensure semantic purity of pseudo samples, we introduce the large-scale multimodal model \textbf{CogVLM} to perform automatic image-text consistency evaluation.

\subsubsection{Action Categories and Prompt Design}
\begin{table}[t]
  \centering
  \caption{Driver Distraction Behavior Categories and Corresponding Prompts}
  \label{tab:prompts}
  \renewcommand{\arraystretch}{1.8}  
  \setlength{\tabcolsep}{4pt}        
  \begin{tabular}{%
    | >{\centering\arraybackslash}m{1.0cm}  
    | >{\centering\arraybackslash}m{3.0cm}  
    | m{4.0cm}  
    |}
    \hline
    \textbf{Category} & \textbf{Behavior Description} & \textbf{Prompt} \\ \hline
    C0 & Normal driving               & The driver is driving normally with both hands on the steering wheel. \\ \hline
    C1 & Texting with right hand       & The driver is texting with the right hand while driving. \\ \hline
    C2 & Holding phone to right ear    & The driver is holding a phone to the right ear while driving. \\ \hline
    C3 & Texting with left hand      & The driver is texting with the left hand while driving. \\ \hline
    C4 & Holding phone to left ear   & The driver is holding a phone to the left ear while driving. \\ \hline
    C5 & Adjusting multimedia         & The driver is adjusting the car's multimedia or infotainment system. \\ \hline
    C6 & Drinking water               & The driver is drinking water while driving. \\ \hline
    C7 & Reaching toward back seat    & The driver is reaching toward the back seat to grab something. \\ \hline
    C8 & Applying makeup              & The driver is applying makeup while driving. \\ \hline
    C9 & Talking to passenger         & The driver is talking to a passenger while driving. \\ \hline
  \end{tabular}
\end{table}

Considering the diversity of distracted driving behaviors, we divide the action space into 10 distinct categories (C0–C9) based on dataset definitions. These categories cover typical activities such as normal driving, calling or texting with either hand, adjusting multimedia systems, drinking, applying makeup, reaching for rear-seat items, and talking to passengers.

For each category, we construct a concise and precise English prompt that guides CogVLM in determining image-text alignment. The prompts are declarative and well-structured, designed to enhance instruction-following performance. Table~\ref{tab:prompts} lists all categories and their corresponding prompts.

\subsubsection{Consistency Scoring Function}

To avoid uncontrollability caused by discrete "yes/no" outputs, this paper designs the following numerical query template to force \texttt{CogVLM} to return a matching score in the interval \([0,1]\):

\begin{quote}
\texttt{How well does this image match the description: “\{Prompt[c]\}”?} \\
\texttt{Respond with a number between 0 and 1, where 1 means perfect match.}
\end{quote}

The image-text matching function is defined as
\[
s = C(I_g, \text{Prompt}[c]), \quad s \in [0,1],
\]

where \(C(\cdot)\) denotes the multimodal similarity mapping of \texttt{CogVLM} after instruction tuning; \(I_g\) is the generated pseudo sample; \text{Prompt}[c] represents the text prompt corresponding to the Category (C0,C1,…,C9). the closer \(s\) is to 1, the higher the semantic consistency.

\subsubsection{Filtering Algorithm and Threshold Setting}
To balance data purity and retention rate, this paper sets a fixed threshold 
\[
\tau = 0.8.
\]

The filtering process is presented in Algorithm~\ref{alg:cogvlm_filter}.
\begin{algorithm}[t]
\caption{CogVLM-based Semantic Filtering}
\label{alg:cogvlm_filter}
\begin{algorithmic}[1]
\Require Pseudo-sample set $\mathbb{S} = \{(I_g, c)\}$, prompt dictionary $\texttt{Prompt}[\cdot]$, threshold $\tau$
\Ensure Filtered pseudo-sample set $\mathbb{S}_{\text{filtered}}$
\State $\mathbb{S}_{\text{filtered}} \gets \emptyset$
\ForAll{$(I_g, c)$ in $\mathbb{S}$}
    \State $\texttt{query} \gets$ ``How well does this image match the description: "\texttt{Prompt}[c]"? Respond with a number between 0 and 1, where 1 means perfect match.''
    \State $s \gets \texttt{CogVLM}(I_g, \texttt{query})$
    \If{$s \ge \tau$}
        \State $\mathbb{S}_{\text{filtered}} \gets \mathbb{S}_{\text{filtered}} \cup \{(I_g, c)\}$
    \EndIf
\EndFor
\State \Return $\mathbb{S}_{\text{filtered}}$
\end{algorithmic}
\end{algorithm}

The final high-quality pseudo sample set is denoted as
\[
D_{\text{pseudo}}^\star = \{ I_g \mid s \geq \tau \}.
\]

\section{Experiment and Analysis}
To validate the performance improvement of the proposed PQ-DAF framework in data-scarce scenarios, we conduct experiments using ResNet50\cite{he2016deep} as the classification model in PQ-DAF, and perform comparative evaluations on the StateFarm and AUC-DDD datasets under both 10-shot and 30-shot settings.

\subsection{Datasets}
\textbf{\emph{StateFarm}}
The StateFarm dataset  is a benchmark for distracted driver detection, released by the Kaggle competition. It includes 10 distraction categories (e.g., texting with right hand, calling with left hand, operating radio, normal driving) with 17,462 labeled RGB images captured by dashboard cameras. The dataset involves 26 subjects of diverse ethnicities, skin tones, and genders, ensuring good diversity. For few-shot experiments, we use the original training set (80\% of data) to fine-tune the PCDMS module of PQ-DAF, and randomly sample the original test set (20\% of data) to construct 10-shot and 30-shot subsets (10/30 images per class) as few-shot training data.

\textbf{\emph{AUC-DDD}}
The AUC-DDD (American University in Cairo Distracted Driver Dataset)  was constructed by Abouelnaga et al. from the American University in Cairo, widely used for distracted driver detection. It includes 10 categories (e.g., safe driving, left-hand calling, right-hand texting, eating, grooming, radio adjustment) with 14,478 images captured in real vehicles from 31 participants across seven countries, covering diverse ethnicities, skin tones, and genders. For few-shot experiments, we use the original training set (80\% of data) to fine-tune the PCDMS module of PQ-DAF, and randomly sample the original test set (20\% of data) to construct 10-shot and 30-shot subsets (10/30 images per class) as few-shot training data.

\subsection{Evaluation Metrics}  
We use Top-1 Accuracy as the evaluation metric to measure model performance, defined as:  
$$ \text{Top-1 Accuracy} = \frac{1}{N} \sum_{i=1}^{N} \mathbb{I}(y_i = \hat{y}_i) $$ 
where $N$ denotes the total number of test samples, $y_i$ is the ground-truth label of sample $i$, $\hat{y}_i$ represents the predicted label by the model, and $\mathbb{I}(\cdot)$ is the indicator function (returning 1 if the prediction is correct, 0 otherwise). 

We use \textit{F1-Score} as an additional evaluation metric to measure model performance, defined as:
$$ \text{F1-Score} = 2 \times \frac{\text{Precision} \times \text{Recall}}{\text{Precision} + \text{Recall}}, $$
where
$$ \text{Precision} = \frac{\text{TP}}{\text{TP} + \text{FP}}, \quad \text{Recall} = \frac{\text{TP}}{\text{TP} + \text{FN}}, $$
$\text{TP}$ denotes true positives, $\text{FP}$ represents false positives, and $\text{FN}$ stands for false negatives.
\subsection{Implementation Details} 

All experiments were conducted on a machine equipped with an NVIDIA GeForce RTX 4070 Super GPU and an Intel Core i5-13600KF CPU. For the PCDMs used in sample generation, input images were uniformly resized to $512 \times 512$. The Adam optimizer was employed with a learning rate (lr) of $5 \times 10^{-6}$ and a weight decay of $1 \times 10^{-3}$. Training was carried out for 30,000 epochs with a batch size of 5. To reduce memory usage, the CogVLM model used for quality validation was loaded using 4-bit quantization.

For all other few-shot learning experiments, the AdamW optimizer was adopted with an lr of $1 \times 10^{-4}$ and a weight decay of $1 \times 10^{-2}$. Each model was trained for 20 epochs with a batch size of 16 to ensure consistency and fairness across experimental settings.

\subsection{Comparison with State-of-the-art Methods} 
\subsubsection{Comparisons on StateFarm}

To further validate the effectiveness of the proposed pseudo-sample augmentation method in few-shot driver behavior recognition tasks, we conducted 10-shot and 30-shot classification experiments on the StateFarm dataset, corresponding to extremely low and low sample conditions in real-world scenarios. Tables~\ref{tab:2} present a comparative analysis of Top-1 classification accuracy on both the original training set and the augmented set (aug), across lightweight CNNs, standard CNNs, and Transformer-based models.

\begin{table}[t]
\renewcommand{\arraystretch}{1.1}
\centering
\caption{Performance Comparison on StateFarm Dataset}
\label{tab:2}
\scalebox{1.0}{
\begin{tabular}{@{}l@{\hspace{0.3em}} c@{\hspace{0.8em}} c@{\hspace{0.6em}} c@{\hspace{0.8em}} c@{}} 
\toprule
\multirow{2}{*}{\textbf{Model}} & \multicolumn{2}{c}{\textbf{10-shot}} & \multicolumn{2}{c}{\textbf{30-shot}} \\
\cmidrule(lr){2-3} \cmidrule(lr){4-5}
& Top-1 (\%) & F1 (\%) & Top-1 (\%) & F1 (\%) \\
\midrule
MobileNetv3\textsubscript{\textcolor{blue}{(ICCV'19)}}\cite{howard2019searching} & 16.67 & 16.17 & 21.33 & 20.71 \\
MobileViT\textsubscript{\textcolor{blue}{(ICLR'22)}}\cite{mehta2021mobilevit} & 17.33 & 16.81 & 31.33 & 30.48 \\
FastViT\textsubscript{\textcolor{blue}{(ICLR'22)}}\cite{vasu2023fastvit} & 24.67 & 23.93 & 48.67 & 46.91 \\
ConvNeXt\textsubscript{\textcolor{blue}{(CVPR'22)}}\cite{liu2022convnet} & 25.33 & 24.57 & 42.67 & 41.12 \\
Inceptionv4\textsubscript{\textcolor{blue}{(AAAI'17)}}\cite{szegedy2017inception} & 26.67 & 25.87  & 60.00 & 58.24 \\
PVT\textsubscript{\textcolor{blue}{(ICCV'21)}}\cite{wang2021pyramid} & 26.67 & 25.87 & 54.00 & 52.17 \\
ResNet50\textsubscript{\textcolor{blue}{(CVPR'16)}}\cite{he2016deep} & 36.67 & 35.57 & 64.67 & 62.38\\
\midrule
\textbf{\textsc{Ours}} & \textbf{54.00} & \textbf{52.38} &\textbf{88.00} & \textbf{85.44} \\
\bottomrule
\end{tabular}
}
\vspace{-0.5em}
\end{table}

As shown in Table~\ref{tab:2}, our proposed method achieves significant advantages over all compared models on the StateFarm dataset under both 10-shot and 30-shot training settings, fully demonstrating its effectiveness and generalization capability under data-scarce conditions. Specifically, in the 10-shot setting, our method attains a Top-1 accuracy of 54.00\%, representing an improvement of 17.33 percentage points over the best baseline model, ResNet50 (36.67\%). In the 30-shot setting, the accuracy further increases to 88.00\%, substantially outperforming the second-best model, Inceptionv4 (60.00\%), and also achieving an F1 score of 85.44\% compared to 62.38\% for ResNet50. This notable improvement can be attributed to the synergistic effect of the pose-guided image generation strategy and the semantic consistency filtering mechanism, which not only enhances the diversity of training samples but also ensures the high quality of synthetic data, enabling the model to learn more robust and discriminative feature representations under limited data conditions.

\subsubsection{Comparisons on AUC-DDD}
Building upon the validation of our model’s effectiveness on the StateFarm dataset, we further evaluate our approach on the AucDDD dataset to assess its generalization ability under different data distributions. The AucDDD dataset contains more diverse driving scenarios and behavior patterns, offering a complementary perspective for analyzing the effectiveness of our data augmentation method.
\begin{table}[t]
\renewcommand{\arraystretch}{1.1}
\centering
\caption{Performance Comparison on AUC-DDD Dataset }
\label{tab:4}
\scalebox{1.0}{
\begin{tabular}{@{}l@{\hspace{0.3em}} c@{\hspace{0.8em}} c@{\hspace{0.6em}} c@{\hspace{0.8em}} c@{}} 
\toprule
\multirow{2}{*}{\textbf{Model}}  & \multicolumn{2}{c}{\textbf{10-shot }} & \multicolumn{2}{c}{\textbf{30-shot}} \\
\cmidrule(lr){2-3} \cmidrule(lr){4-5}
& Top-1 (\%) & F1 (\%) & Top-1 (\%) & F1 (\%) \\ 
\midrule
MobileNetv3\textsubscript{\textcolor{blue}{(ICCV'19)}}\cite{howard2019searching}  & 18.67 & 16.49 & 31.33 & 31.00 \\
Inceptionv4\textsubscript{\textcolor{blue}{(AAAI'17)}}\cite{szegedy2017inception} & 19.33 & 18.65  & 39.19 & 39.40 \\
ConvNeXt\textsubscript{\textcolor{blue}{(CVPR'22)}}\cite{liu2022convnet}  & 21.33 & 20.90 & 40.54 & 39.83 \\
PVT\textsubscript{\textcolor{blue}{(ICCV'21)}}\cite{wang2021pyramid} & 22.67 & 22.74 & 46.67 & 46.10 \\
MobileViT\textsubscript{\textcolor{blue}{(ICLR'22)}}\cite{mehta2021mobilevit} & 23.33 &23.98 &37.16 & 36.20 \\
FastViT\textsubscript{\textcolor{blue}{(ICLR'22)}}\cite{vasu2023fastvit}  & 24.67 & 25.58 & 42.67& 41.87 \\
ResNet50\textsubscript{\textcolor{blue}{(CVPR'16)}}\cite{he2016deep} & 30.67 & 31.08 & 50.67 & 50.68\\
\midrule
\textbf{\textsc{Ours}}  & \textbf{40.67} & \textbf{38.47} &\textbf{71.33} & \textbf{71.27} \\
\bottomrule
\end{tabular}
}
\vspace{-0.5em}
\end{table}

As shown in Table~\ref{tab:4}, on the more challenging AUC-DDD dataset, the proposed method achieves the best results across different training data scales, demonstrating outstanding robustness. In the 10-shot setting, our method attains a Top-1 accuracy of 40.67\%, representing a 10.00 percentage point improvement over the best-performing baseline, ResNet50 (30.67\%). In the 30-shot setting, the accuracy further climbs to 71.33\%, far exceeding the second-best model, PVT (46.67\%), and achieving an F1 score of 71.27\%, which is substantially higher than ResNet50’s 50.68\%. These results further validate the effectiveness of the proposed approach on the AUC-DDD dataset and are consistent with the performance trends observed on the StateFarm dataset.

\subsubsection{Data Augmentation Comparison}

To rigorously assess the effectiveness of the proposed PQ-DAF in enhancing few-shot driver behavior recognition, we conducted a comparative evaluation using ResNet-50 as the baseline model. This evaluation includes seven widely adopted data augmentation techniques—AugMix, Manifold, Mixup, CutMix, PixMix, PuzzleMix, and GuidedMixup—encompassing a diverse range of strategies such as pixel-level blending, semantic-guided mixing, and robustness-oriented transformations. Top-1 classification accuracy is adopted as the primary metric to quantitatively benchmark the performance across methods.
\begin{table}[t]
\renewcommand{\arraystretch}{1.1}
\centering
\caption{Top-1 Accuracy (\%) Comparison of Different Data Augmentation Methods}
\label{tab:6}
\scalebox{1.1}{ 
\begin{tabular}{@{}l@{\hspace{0.8em}} c@{\hspace{1.2em}} c@{}} 
\toprule
\textbf{Method} & \multicolumn{1}{c}{\textbf{Statefarm}} & \multicolumn{1}{c}{\textbf{AUC-DDD}} \\
                & \multicolumn{1}{c}{\textbf{Top-1 Acc. (\%)}} & \multicolumn{1}{c}{\textbf{Top-1 Acc. (\%)}} \\
\midrule
AugMix \cite{hendrycks2019augmix} & 40.67 ± 2.34 & 33.44 ± 1.82 \\
Manifold \cite{verma2019manifold} & 39.46 ± 0.37 & 32.56 ± 1.60 \\
Mixup \cite{zhang2017mixup} & 39.67 ± 1.90 & 32.96 ± 1.73 \\
CutMix \cite{yun2019cutmix} & 39.78 ± 0.94 & 33.97 ± 1.45 \\
PixMix \cite{hendrycks2022pixmix} & 33.44 ± 3.00 & 29.10 ± 1.25 \\
PuzzleMix \cite{kim2020puzzle} & 41.37 ± 0.82 & 34.42 ± 1.28\\
GuidedMixup \cite{kang2023guidedmixup} & 42.45 ± 0.54 & 35.63 ± 1.17 \\
\midrule
\textbf{\textsc{Ours}} & \textbf{54.67 ± 1.23} & \textbf{40.67 ± 1.12} \\
\bottomrule
\end{tabular}
}
\vspace{-0.5em}
\end{table}
\begin{figure*}[t]
    \centering
    \includegraphics[width=1\linewidth]{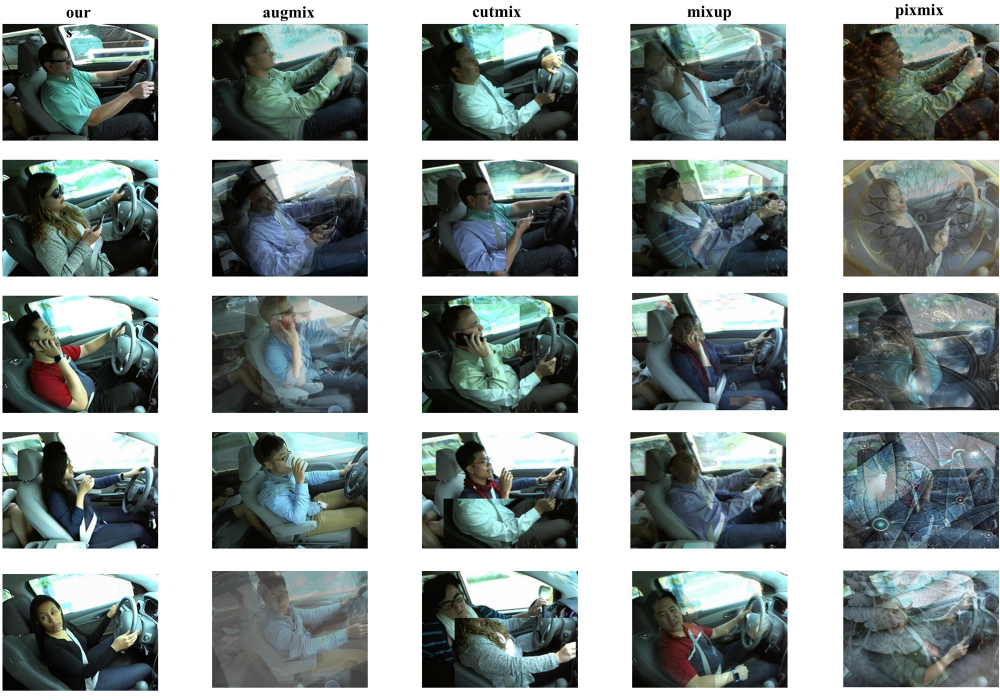}
\caption{Visual Comparison of Data Augmentation Techniques }
\label{fig:aug}
\end{figure*}

\begin{figure}[t]
    \centering
    \includegraphics[width=1\linewidth]{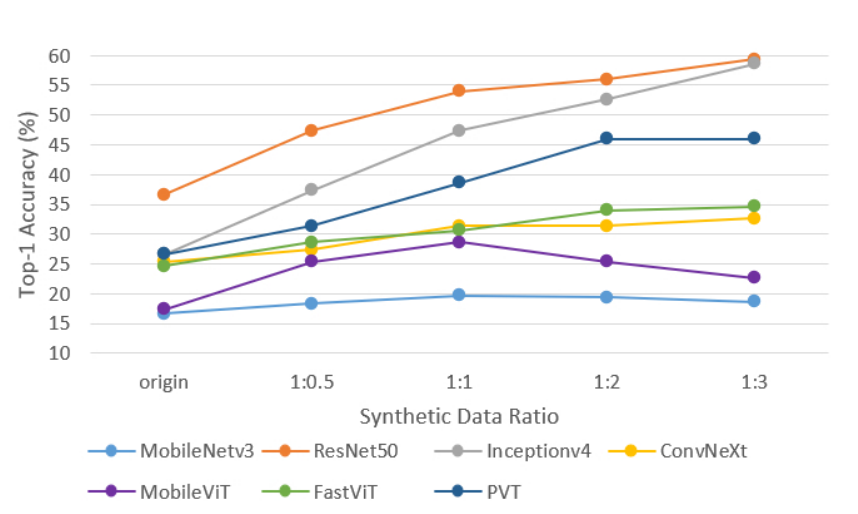}
\caption{Accuracy Comparison Under Different Mixture Ratios of Real and Synthetic Data }
\label{fig:4}
\end{figure}

\begin{table}[t]
\renewcommand{\arraystretch}{1.1}
\centering
\caption{Accuracy Comparison Under Different Mixture Ratios of Real and Synthetic Data (10-Shot Setting)}
\label{tab:7}
\scalebox{0.8}{
\begin{tabular}{@{}l c c c c c c@{}} 
\toprule
\textbf{Model} & \textbf{Params (M)} & \textbf{Real Only} & \textbf{1:0.5} & \textbf{1:1} & \textbf{1:2} & \textbf{1:3} \\
\midrule
MobileViT\textsubscript{\textcolor{blue}{(ICLR'22)}}\cite{mehta2021mobilevit} & 0.95 & 17.33 & 25.33 & 28.67 & 25.33 & 22.67 \\
MobileNetv3\textsubscript{\textcolor{blue}{(ICCV'19)}}\cite{howard2019searching} & 1.53 & 16.67 & 17.33 & 18.67 & 19.33 & 18.67 \\
FastViT\textsubscript{\textcolor{blue}{(ICLR'22)}}\cite{vasu2023fastvit} & 3.24 & 24.67 & 28.67 & 30.67 & 34.00 & 34.67 \\
PVT\textsubscript{\textcolor{blue}{(ICCV'21)}}\cite{wang2021pyramid} & 3.41 & 26.67 & 31.33 & 38.67 & 46.00 & 46.00 \\
ConvNeXt\textsubscript{\textcolor{blue}{(CVPR'22)}}\cite{liu2022convnet} & 4.83 & 25.33 & 27.33 & 31.33 & 31.33 & 32.67 \\
ResNet50\textsubscript{\textcolor{blue}{(CVPR'16)}}\cite{he2016deep} & 25.56 & 36.67 & 47.33 & 54.00 & 56.00 & 59.33 \\
Inceptionv4\textsubscript{\textcolor{blue}{(AAAI'17)}}\cite{szegedy2017inception} & 41.10 & 26.67 & 37.33 & 47.33 & 52.67 & 58.67 \\
\bottomrule
\end{tabular}
}
\vspace{0.3em}
\begin{tablenotes}
\footnotesize
\item \textit{Note}: Real Only indicates using only real samples; 1:x indicates the ratio of real to generated data.
\end{tablenotes}
\end{table}

\begin{figure*}[t]
    \centering
    \includegraphics[width=1\linewidth]{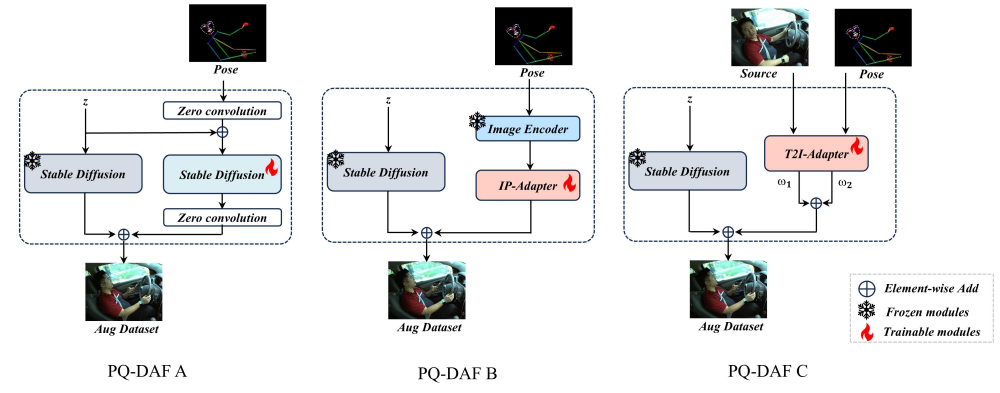}
\caption{Generation Model Variants of the PQ-ADF Framework with All Other Processes Identical}
\label{fig:5}
\end{figure*}

\begin{figure*}[t]
    \centering
    \includegraphics[width=1\linewidth]{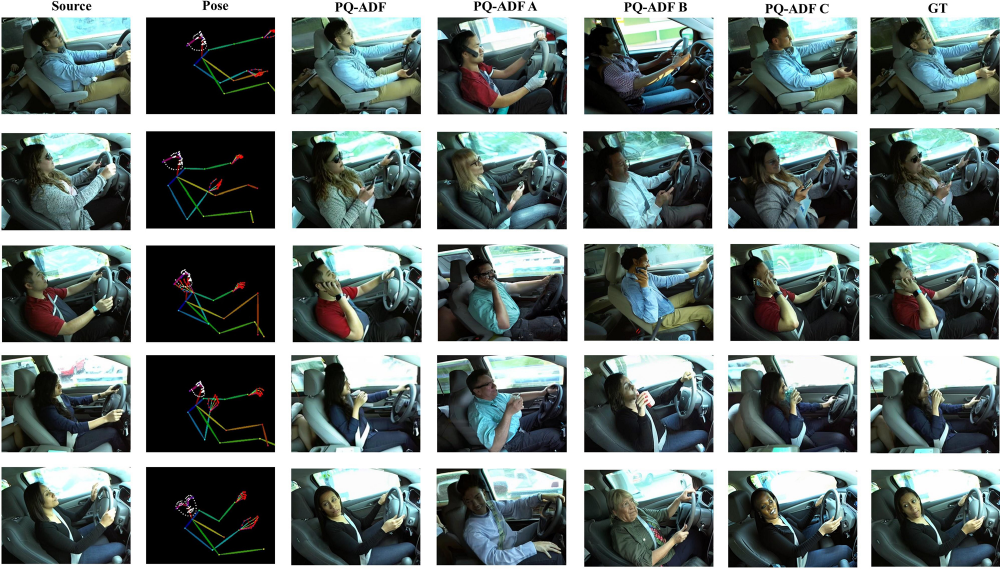}
\caption{ Visual Comparisons with three variants.}
\label{fig:xr}
\end{figure*}
Table~\ref{tab:6} shows that our proposed data augmentation method significantly outperforms several existing mainstream augmentation strategies on both the StateFarm and AUC-DDD datasets. For example, on the StateFarm dataset, our method achieves a Top-1 accuracy of 54.67\%, which is more than 12 percentage points higher than the second-best method, GuidedMixup (42.45\%). On the more challenging AUC-DDD dataset, our approach also leads with a Top-1 accuracy of 40.67\%, surpassing all other augmentation techniques. Notably, conventional synthetic augmentation methods such as Mixup, CutMix, and PuzzleMix generally achieve around 30\% accuracy on both datasets, indicating a significant performance gap. These results demonstrate that our method is more effective in enhancing model discrimination and generalization performance. This may be attributed to the fact that the generated pseudo-samples by our approach possess higher semantic consistency and contextual integrity, enabling the model to better exploit latent information under few-shot learning conditions and thus achieve more robust performance.

In addition, we provide visualizations of the augmented images generated by all methods except PuzzleMix and GuidedMixup (as these are performed in the latent space). As shown in Fig. \ref{fig:aug}, our method achieves the best performance in preserving behavioral semantics, clarity, and naturalness. Compared to methods such as AugMix and Mixup, our augmented samples maintain the coherence of human actions and scene structure, avoiding mixed artifacts and semantic ambiguity. Meanwhile, in contrast to the unnatural occlusion or overly stylized features produced by CutMix and PixMix, our method facilitates more accurate capture of key behavioral cues by the model.
\subsubsection{Impact of Generated Data Ratios on Model Performance}
This experiment aims to evaluate the impact of mixing real and synthetic data in extremely low-data scenarios, where training samples are highly limited, and to verify the effectiveness of synthetic data generated by the proposed PQE-AD framework in enhancing model performance. Specifically, under a 10-sample-per-class setting, we compare the classification accuracy of seven mainstream models—MobileNetv3, ResNet50, Inceptionv4, ConvNeXt, MobileViT, FastViT, and PVT—across different real-to-synthetic data ratios (1:0.5, 1:1, 1:2, 1:3).


As shown in Table \ref{tab:7} and Fig. \ref{fig:4}, under the 10-shot setting, models of different scales all exhibit varying degrees of performance improvement when trained with different mixture ratios of real and synthetic data, but the extent of improvement is closely tied to model capacity. Overall, as the proportion of synthetic data increases, most models achieve steady gains in classification accuracy, with larger-capacity models benefiting the most. For example, ResNet50 (25.56M parameters) achieves 47.33\% accuracy at a real-to-synthetic ratio of 1:0.5, which rises to 59.33\% at 1:3—an increase of 12 percentage points. Similarly, Inceptionv4 (41.10M parameters) improves from 37.33\% to 58.67\%, an increase of more than 21 percentage points, demonstrating the strong potential of high-capacity models in leveraging high-quality synthetic samples.

In contrast, lightweight models show more limited gains and are more sensitive to the mixture ratio. For instance, MobileNetv3 (1.53M parameters) reaches its peak performance at a 1:2 ratio but with only a small improvement, while MobileViT (0.95M parameters) even suffers performance degradation when the ratio becomes too high. This suggests that for lightweight networks, a moderate mixture ratio (recommended between 1:1 and 1:2) is more suitable, as it balances sample diversity with the preservation of the original distribution. Excessive proportions of synthetic data may introduce noise or domain bias that negatively impacts performance due to limited model capacity. Therefore, in practice, it is advisable to adjust the ratio according to the specific model: for medium-to-large capacity models, gradually increasing the proportion of synthetic samples can yield greater benefits, whereas for lightweight models, adopting a conservative ratio of 1:1 or 1:2 is preferable for stable performance gains. This experiment validates the effectiveness of PQE-AD in generating useful data under extreme low-data conditions and highlights the differences in optimal mixture strategies for models of varying capacities in real-world deployments.

\subsection{Ablation Studies and Analysis}  

To validate the effectiveness of the proposed PQ-ADF framework in selecting the PCDMs module, we introduced different generation models into the PQ-ADF framework to construct three variants, as shown in Fig. \ref{fig:5}. Specifically, PQ-ADF A adopts ControlNet \cite{zhang2023adding} as the generation model, using the extracted pose information as the conditional input to guide the generation process. While this approach can control the pose of generated images to some extent, it cannot constrain identity features, resulting in random variations in the generated person’s appearance. PQ-ADF B employs IP-Adapter \cite{ye2023ip} as the generation model, which similarly uses only the pose image as the conditional input, and thus shares the same limitation of not controlling person-specific features. PQ-ADF C utilizes T2I-Adapter \cite{mou2024t2i} as the generation model. This method supports flexible multi-condition control, and during generation, both the identity features and pose features are provided as conditional inputs, enabling more precise alignment between the person’s identity and pose.

\begin{table}[t]
\renewcommand{\arraystretch}{1.1}
\centering
\caption{Accuracy Comparison of Generative Models Using ResNet-50 Backbone}
\label{tab:8}
\scalebox{1.10}{ 
\begin{tabular}{@{}l c@{}} 
\toprule
\textbf{Method} & \textbf{Top-1 Accuracy (\%)} \\ 
\midrule
PQ-ADF A\textsubscript & 45.67 ± 1.57 \\

PQ-ADF B\textsubscript &  43.78 ± 2.31\\
PQ-ADF C\textsubscript &  49.92 ± 1.31\\ 
\midrule 
\textbf{\textsc{PQ-ADF}} & \textbf{54.67 ± 1.23} \\ 
\bottomrule
\end{tabular}
}
\vspace{0.0em}
\begin{tablenotes}
\footnotesize
\item \textit{Note}: All models are evaluated on the same driver distraction dataset with 10-shot training.
\end{tablenotes}
\end{table}

As shown in Table~\ref{tab:8}, the samples generated using the proposed PCDMs achieved the best performance improvement, with an accuracy of 54.67\%, significantly outperforming other generation methods. These results demonstrate that the PCDMs module, through multi-level feature alignment, is better suited for the driver distraction detection task, producing driver samples whose appearance and actions are highly consistent with real data.

Furthermore, we provide visual comparisons of the samples generated by each variant. As shown in the Fig. \ref{fig:xr}, the samples generated by our model appear more visually consistent with real data. PQ-ADF C ranks second in terms of realism, while PQ-ADF A and PQ-ADF B produce lower-quality outputs due to their reliance on pose-only conditioning, resulting in identity inconsistency and less realistic samples.

\section{Conclusion}
This study proposed a Pose-driven Quality-controlled Data Augmentation Framework (PQ-DAF) to address performance degradation in few-shot driver distraction detection caused by limited annotations and domain shifts. The framework employed a Progressive Conditional Diffusion Model to synthesize pose-consistent, diverse samples and leveraged a vision-language model for semantic quality filtering, effectively enhancing data-scarce generalization. Experiments showed substantial performance gains, including an improvement of ResNet50 Top-1 accuracy from 36.67\% to 54.00\% on StateFarm and Inceptionv4 from 19.33\% to 34.00\% on AUC-DDD. Nevertheless, the approach remained dependent on pose estimation accuracy and filtering robustness. Future work will explore advanced pose estimation and multi-modal scene understanding to further improve semantic fidelity, optimize sample selection, and extend applicability to more complex driving scenarios and behavior categories.


\bibliographystyle{IEEEtran}
\bibliography{ref}

\end{document}